\documentclass[runningheads]{llncs}

\usepackage{eccv}

\usepackage{eccvabbrv}
\usepackage{graphicx}
\usepackage{booktabs}
\usepackage[accsupp]{axessibility}  
\usepackage{orcidlink}
\usepackage{tabularx}

\newcommand{\tc}[1]{\tiny\cite{#1}}

\begin{document}

\title{SEMIR: Topology-Preserving Graph Minors for Thin-Structure Segmentation} 
\titlerunning{SEMIR: Graph Minors for Thin-Structure Segmentation}

\author{Luke James Miller\inst{1}\orcidlink{0009-0001-4612-5402} \and
Yugyung Lee\inst{1}\orcidlink{0000-0002-1619-1695}}
\authorrunning{L.~Miller and Y.~Lee}

\institute{University of Missouri-Kansas City, Kansas City MO 64110, USA
\email{ljmbm5@umsystem.edu}}

\maketitle
\begin{center}
\end{center}

\begin{abstract}
Thin-structure segmentation—power lines, cracks, lane markings at 1–3 pixel width—requires preserving connectivity that standard representations preclude: patching severs continuous structures and conventional superpixels merge thin targets into background before classification. Topology-aware losses penalize connectivity breaks at the objective level but cannot recover what the representation has already destroyed. We propose SEMIR, a framework that replaces the pixel lattice with a parameterized graph minor whose contraction map preserves thin-structure connectivity under the contraction criterion. The minor collapses millions of pixels into tens or hundreds of boundary-aligned supernodes, enabling full-resolution inference without patching at scales demonstrated up to 21 MP in this paper; a lightweight GNN classifies the reduced graph and an exact map lifts predictions to pixel resolution. One pipeline—identical architecture, features, loss, and GNN hyperparameters across all datasets—matches or exceeds domain-specific baselines on TTPLA (power lines), CrackSeg9k (pavement cracks), and SkyScapes Lane (aerial markings) on Dice, IoU, and Boundary F1 while reducing mask fragmentation by at least 4.6$\times$ relative to SLIC at matched inference.
\end{abstract}

\section{Introduction}\label{sec:intro}

Thin structures---power lines, pavement cracks, lane markings---pervade visual scenes across imaging domains, spanning one to three pixels in width yet carrying connectivity as their primary semantic signal: a fragmented power line is a missed hazard, a disconnected crack network mischaracterizes structural integrity, a broken lane marking disrupts route topology. Standard segmentation pipelines process these targets on the native pixel grid or on task-agnostic superpixel decompositions (SLIC~\cite{achanta2012slic}, Felzenszwalb--Huttenlocher~\cite{felzenszwalb2004efficient}) that agglomerate without regard to target geometry, fragmenting thin structures before any classifier operates. Once a two-pixel conductor is split across adjacent superpixels, no downstream refinement recovers the discarded connectivity.

Recent work addresses this failure through the training objective. Topology-aware losses---clDice~\cite{shit2021cldice}, Skeleton Recall Loss~\cite{skeleton_recall_eccv2024}, and Topograph~\cite{topograph2024}---penalize connectivity breaks via skeleton- or persistence-based differentiable surrogates. These methods target the correct failure mode but operate on a representation with no structural obligation to preserve connectivity: the pixel grid encodes spatial adjacency uniformly without distinguishing thin targets from bulk regions, and superpixel methods actively merge across thin-structure boundaries. Topology preservation in such pipelines is an emergent property of optimization, not a formulation of the representation.

We attack the complementary axis. SEMIR constructs a parameterized graph minor from the pixel lattice via edge contraction, node deletion, and edge deletion, replacing the grid with a compact, boundary-aligned graph on which a GNN performs node-level classification. The critical property is graph-theoretic: edge contraction defines a surjective map from the parent graph $G$ to the minor $H$ such that any connected subgraph of $G$ whose edges are all contracted---rather than deleted---maps to a connected subgraph of $H$~\cite{robertson2004graph,diestel2017graph}. For thin structures, this surjection has a direct consequence: a power line or crack whose pixels are intensity-homogeneous will have every internal edge contracted into a single supernode chain, preserving the structure's connectivity in $H$ regardless of its width or orientation. The constraint is structural, not statistical---it holds for any input satisfying the contraction criterion, independent of training data or loss landscape.

Minor parameters are set via few-shot boundary-alignment optimization on 5--20 labeled images; a GINE classifier~\cite{hu2020strategies} operates on the reduced graph; an exact lifting map, defined by the partition induced during contraction, recovers pixel-level predictions without interpolation. The entire pipeline is domain-agnostic: the same architecture, features, loss, and hyperparameters are applied to every dataset without modification. This work extends our earlier graph-minor representation framework~\cite{miller2026semir} to thin-structure segmentation domains.

We evaluate SEMIR on three benchmarks spanning distinct thin-structure domains: TTPLA~\cite{abdelfattah2022ttpla} (power lines, 1100 UAV images, $3840\times2160$), CrackSeg9k~\cite{kulkarni2022crackseg9k} (pavement cracks, 9255 images, $400\times400$), and SkyScapes~\cite{azimi2019skyscapes} (aerial lane markings, 16 images, $5616\times3744$, 13\,cm/pixel GSD). These datasets span two orders of magnitude in image resolution, three orders in dataset size, and cover thin-structure morphologies from near-linear to irregularly branching to discrete and periodic.

Our contributions:
\begin{enumerate}
    \item We show that graph-minor regionization provides a formal connectivity constraint for thin structures: any connected foreground region whose internal edges satisfy the contraction criterion is preserved as a connected subgraph in the minor, independent of topology-aware losses or task-specific architecture.
    \item We evaluate SEMIR across three thin-structure domains with a single pipeline, matching or exceeding domain-specific baselines on Dice, IoU, and Boundary F1 while achieving consistently lower mask fragmentation.
    \item We isolate the source of improvement via controlled ablation replacing the minor with SLIC superpixels at matched region count, confirming that the topological advantage originates in the representation, not in region-level inference alone.
\end{enumerate}

\section{Related Work}\label{sec:related}

\paragraph{Thin-structure segmentation.}\label{sec:related:thin}
Domain-specific architectures dominate thin-structure benchmarks. For power-line detection, dedicated encoder--decoder designs~\cite{abdelfattah2022ttpla} and line-prior networks~\cite{plrefiner,duformer} exploit geometric regularity in aerial imagery. Crack segmentation methods~\cite{hrsegnet,crackformer2} integrate multi-scale fusion and boundary-aware attention to resolve hairline fractures against cluttered backgrounds. Aerial lane-marking extraction~\cite{azimi2019skyscapes} adapts general-purpose backbones---Swin~\cite{liu2021swin}, SegFormer~\cite{xie2021segformer}, DeepLabV3+~\cite{chen2018deeplabv3plus}---to overhead imagery, relying on high-resolution input and class-balanced sampling to compensate for extreme foreground--background imbalance. In every case the representation is the pixel grid or a fixed multi-scale hierarchy derived from it; none incorporates a structural constraint against thin-structure fragmentation. SEMIR replaces the grid with a topology-preserving graph minor, making it orthogonal to these architectural choices and applicable across all three domains with a single pipeline.

\paragraph{Topology-aware losses.}\label{sec:related:topology}
clDice~\cite{shit2021cldice} differentiates through soft skeletonization to penalize connectivity breaks in tubular structures. Skeleton Recall Loss~\cite{skeleton_recall_eccv2024} improves computational efficiency by restricting the topological penalty to recall on skeletonized ground truth. Topograph~\cite{topograph2024} constructs a graph from the prediction and penalizes discrepancies in connected-component counts via a differentiable Betti-number surrogate. Stucki \etal~\cite{stucki2023topologically} propose a topologically faithful loss using persistent homology; Lux \etal~\cite{lux2024topological} extend this to multi-class settings. Morphological post-processing (dilation, erosion, skeletonization) provides a complementary family of deterministic topology repair operators~\cite{serra1983morphology}. All operate on the pixel grid: they modify the training objective but leave the representation unchanged. SEMIR is complementary---it constrains connectivity at the representation level and could in principle be combined with any topology-aware loss, though we do not test such combinations in this work.

\paragraph{Superpixels and GNN pooling.}\label{sec:related:superpixel}
SLIC~\cite{achanta2012slic} and Felzenszwalb--Huttenlocher~\cite{felzenszwalb2004efficient} reduce inference units via task-agnostic agglomeration with manually tuned parameters. Learned superpixel networks (SSN~\cite{jampani2018ssn}, SEAL~\cite{tu2018seal}) improve boundary adherence but remain grid-bound and provide no formal relationship between induced regions and the original image. Differentiable graph pooling---DiffPool~\cite{ying2018diffpool}, MinCutPool~\cite{bianchi2020mincutpool}, DMoN~\cite{tsitsulin2023dmon}---learns soft cluster assignments for graph coarsening but couples pooling to the task-specific loss, lacks exact lifting, and offers no topology guarantees. SEMIR differs on three axes: graph minor construction is governed by parameterized graph operations with formal connectivity preservation~\cite{robertson2004graph,diestel2017graph, demaine2005algorithmic}; parameters are optimized for boundary alignment via few-shot black-box optimization rather than end-to-end gradient flow; and the lifting map from minor predictions to pixel labels is exact by construction.

\section{Method}\label{sec:method}

\begin{figure}
    \centering
    \includegraphics[width=1\linewidth]{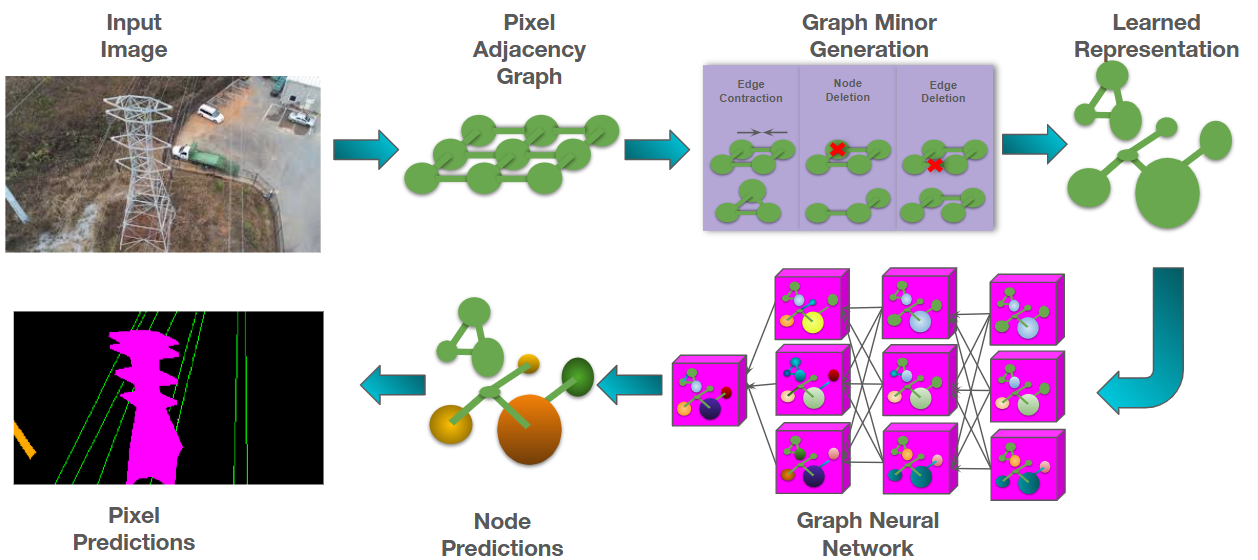}
    \caption{Overview of the SEMIR pipeline. An input image is first encoded as a pixel adjacency graph. A parameterized graph minor is then constructed through three operations—edge contraction, node deletion , and edge deletion—producing a compact, boundary-aligned representation. The surjective structure of contraction enforces that thin-structure connectivity present in the original graph is preserved in the minor. A GINE network performs node classification on the reduced graph using supernode and edge features, and supernode predictions are lifted to pixel resolution via the exact partition encoded during contraction, requiring no interpolation. Minor parameters $\Theta = \{\psi, \alpha, \beta\}$ are set once per dataset through few-shot SMBO on 5--20 labeled images; all other components remain identical across domains.}
    \label{fig:pipe}
\end{figure}
SEMIR takes an RGB image $I \in \mathbb{R}^{H \times W \times C}$ and produces a pixel-level segmentation $\hat{Y} \in \{0,...,k\}^{H \times W}$ for a specified thin-structure target. The pipeline comprises four stages: (i) construction of a pixel-adjacency graph $G$ from $I$; (ii) derivation of a graph minor $H \preceq G$ via parameterized edge contraction, node deletion, and edge deletion; (iii) node-level classification on $H$ by a graph neural network; and (iv) exact lifting of supernode predictions to pixel resolution. Minor parameters are set once per dataset through few-shot boundary-alignment optimization. Figure~\ref{fig:pipe} illustrates the pipeline end to end.

\subsection{Pixel Graph Construction}\label{sec:method:graph}

The image $I$ is encoded as an 8-connected grid graph $G = (V(G), E(G))$ with one node per pixel and edges between all horizontally, vertically, and diagonally adjacent pairs:
\begin{align}
  V(G) &= \{(j,k) : 1 \le j \le H,\; 1 \le k \le W\}, \\
  E(G) &= \{(u,v) \in V(G)^2 : u \neq v,\; \|u - v\|_\infty = 1\}.
\end{align}
 Rather than maintaining explicit coordinate lists or adjacency structures for $G$---which would require $O(|V(G)| + |E(G)|)$ storage, roughly $9HW$ entries for an 8-connected grid---we represent the graph via an expanded binary tensor $T \in \{0,1\}^{(2H{-}1) \times (2W{-}1)}$ that interleaves node and edge states: positions with both indices even encode nodes; positions with at least one odd index encode edges. This representation costs $(2H{-}1)(2W{-}1)$ bits, a fraction of the memory required by a sparse adjacency list, while encoding the full graph topology in a structure that supports $O(|E(G)| - |E(H)|)$ flood-fill construction of the minor via simple array writes. Because $T$ also encodes node/edge states; supernode membership is recovered implicitly from contraction using tensor indices during lifting (no stored adjacency lists). Lifting predictions from $H$ back to pixel resolution reduces to reading $T$ directly---no inverse mapping or interpolation is needed. The modest overhead of the expanded tensor is more than offset by the downstream reduction in data: once the minor is constructed, all subsequent computation operates on $|V(H)|$ supernodes rather than $HW$ pixels.

\subsection{Graph Minor Construction}\label{sec:method:minor}

A graph minor $H \preceq G$ is obtained from $G$ by a sequence of edge contractions, node deletions, and edge deletions~\cite{robertson2004graph,diestel2017graph}. SEMIR parameterizes these operations by a compact parameter set $\Theta = \{\psi, \alpha, \beta\}$ and executes them in a fixed order during a coprime traversal of $V(G)$.

The order in which pixels are visited during contraction affects the resulting minor: a raster-scan traversal preferentially merges rightward and downward, producing elongated supernodes aligned with the scan direction. To eliminate this bias, we traverse the $N = HW$ pixels with a fixed step size $s$ satisfying $\gcd(s, N) = 1$, so the sequence $\{(k \cdot s) \bmod N\}_{k=0}^{N-1}$ visits every pixel exactly once in a deterministic scrambled order. The step $s$ is chosen as the nearest integer to $\lfloor N/2 \rfloor$ that is coprime to $N$. This traversal is deterministic, reproducible, and adds no memory overhead.

\paragraph{Connectivity guarantee.}
Edge contraction defines a surjective homomorphism $\phi: G \to H$ such that for any connected subgraph $S \subseteq G$ whose edges are all contracted (rather than deleted), $\phi(S)$ is a connected subgraph of $H$~\cite{robertson2004graph,diestel2017graph}. For thin structures, this means any foreground region whose pixels are intensity-homogeneous under $\psi$ is preserved as a connected supernode chain in $H$, regardless of width or orientation.

\paragraph{Edge contraction.} Two adjacent supernodes $v_i, v_k \in V(H)$ are merged if their canonical intensity vectors satisfy
\begin{equation}\label{eq:contraction}
  \|{I}_{v_i} - {I}_{v_k}\|_n \le \psi,
\end{equation}
where $\|\cdot\|_n$ is a configurable $L_n$ norm and $\psi \in \Theta$ is the contraction threshold. Contraction recurses until no valid merges remain. 

\paragraph{Node deletion.} Supernodes violating retention criteria are removed:
\begin{equation}\label{eq:node_del}
  V_{\mathrm{del}} := \{v \in V(H) : a_v < \beta_{\min} \;\lor\; a_v > \beta_{\max} \;\lor\; {I}_v \notin [\mathbf{m}_{\min}, \mathbf{m}_{\max}]\},
\end{equation}
where $a_v = |\mathcal{P}_v|$ is the supernode area (pixel count), $I_v$ is the intensity of the canonical pixel of the supernode, and $\beta = (\beta_{\min}, \beta_{\max}, \mathbf{m}_{\min}, \mathbf{m}_{\max}) \subset \Theta$ are retention bounds. ``Node deletion'' means removal from the active prediction graph, not loss of pixel coverage; filtered supernodes are routed to background before exact lifting.

\paragraph{Edge deletion.} Edges across strong intensity gradients are severed:
\begin{equation}\label{eq:edge_del}
  E_{\mathrm{del}} := \{(v_i, v_j) \in E(H) : \|I_{v_i} - I_{v_j}\|_n > \alpha\},
\end{equation}
where $\alpha \in \Theta$ is the deletion threshold. Removed edges partition the minor into components aligned with image boundaries.

\subsection{Few-Shot Parameter Optimization}\label{sec:method:fewshot}

Rather than tuning $\Theta$ manually, we frame parameter selection as black-box optimization over the discrete parameter space. Given a small held-out set $\mathcal{D}_{\mathrm{few}} \subset \mathcal{D}$ of 5--20 labeled images from the training set, we solve
\begin{equation}\label{eq:fewshot}
  \Theta_{\mathrm{opt}} = \arg\min_\Theta \; \mathbb{E}_{(I,Y) \sim \mathcal{D}_{\mathrm{few}}} \bigl[\, 1 - \mathrm{DSC}\bigl(S_B(T, \Theta),\; Y_B\bigr) \,\bigr],
\end{equation}
where $S_B(T, \Theta)$ extracts the boundary elements of the induced minor and $Y_B$ is the ground-truth boundary map derived from the target label map $Y$. Optimization uses sequential model-based optimization (SMBO) with an ExtraTrees surrogate~\cite{hutter2011smbo}. The resulting $\Theta_{\mathrm{opt}}$ defines a family of boundary-aligned partitions adapted to the target structure.

\subsection{Supernode Features}\label{sec:method:features}

Each supernode $u \in V(H)$ aggregates statistics from its constituent pixel set $\mathcal{P}_u$. We compute: area $a_u = |\mathcal{P}_u|$; mean intensity $\bar{I}_u \in \mathbb{R}^3$; per-channel standard deviation $\sigma_u \in \mathbb{R}^3$; boundary length $b_u = |\{(p,q) \in E(G) : p \in \mathcal{P}_u, q \notin \mathcal{P}_u\}|$; compactness $c_u = 4\pi a_u / (b_u^2 + \epsilon)$; elongation $e_u = \sqrt{(\lambda_{u,1}+\epsilon)/(\lambda_{u,2}+\epsilon)}$ where $\lambda_{u,1} \ge \lambda_{u,2} \ge 0$ are eigenvalues of the 2D spatial covariance of pixel coordinates in $\mathcal{P}_u$; and the dominant axis $\mathbf{d}_u \in \mathbb{R}^2$ (eigenvector of $\lambda_{u,1}$). Edge features for $(u,v) \in E(H)$ encode scale-invariant log-ratios of scalar node features, relative displacement, and orientation difference between dominant axes. All edge features are rotation- and scale-robust by construction, and can be used exclusively to remove bias from predictions based on node scale and orientation.

\subsection{GNN Classification and Exact Lifting}\label{sec:method:gnn}

A three-layer GINE network~\cite{hu2020strategies} with relational edge features performs node classification on $H$, producing supernode predictions $\hat{Y}_H \in \{0,\cdots, k\}^{|V(H)|}$. Each graph is processed as a single batch element (batch size one graph per image). The final pixel-level segmentation is obtained by lifting:
\begin{equation}\label{eq:lifting}
  \hat{Y}(j,k) = \hat{Y}_H(u) \quad \text{for all } (j,k) \in \mathcal{P}_u,
\end{equation}
where the partition $\{\mathcal{P}_u\}_{u \in V(H)}$ is encoded in $T$. This lifting is exact: every pixel inherits its supernode's prediction with no interpolation, and the partition is surjective by construction of the contraction. All metrics are computed on $\hat{Y}$ at full pixel resolution.

\section{Experiments}\label{sec:experiments}

\subsection{Datasets}\label{sec:experiments:datasets}

We evaluate on three benchmarks spanning distinct thin-structure domains, imaging modalities, and resolutions (Table~\ref{tab:datasets}). The selection is deliberate: these datasets differ in image scale by two orders of magnitude, in dataset size by three orders, and in thin-structure morphology from near-linear (power lines) to irregularly branching (cracks) to discrete and periodic (dashed lane markings). A method that performs well across all three must generalize over structure geometry, imaging conditions, and foreground--background statistics---the precise claim we make for SEMIR.

\textit{TTPLA}~\cite{abdelfattah2022ttpla} contains 1100 UAV images at $3840\times2160$ resolution depicting overhead power lines. Foreground structures span 1--2 pixels in width against cluttered backgrounds of vegetation, roads, and buildings. Power lines are near-linear over local neighborhoods but vary in orientation, sag, and overlap across the image, making them a test case for connectivity preservation over long spatial extents. The extreme foreground--background imbalance---power-line pixels constitute well under 1\% of the image area---means that even small fragmentation errors translate to large relative drops in overlap metrics. We use the official train/test split with 5-class annotations collapsed to binary (power line vs.\ background).

\textit{CrackSeg9k}~\cite{kulkarni2022crackseg9k} comprises 9255 pavement images at $400\times400$, each with binary pixel-level crack annotations. Cracks range from 1--3 pixels wide and exhibit branching, irregular curvature, and low contrast against asphalt. Unlike the near-linear geometry of power lines, cracks form tree-like networks with T- and Y-junctions whose topology carries direct engineering significance: the number and connectivity of branches determines a crack network's severity classification. CrackSeg9k aggregates images from multiple sources and surface types, introducing variation in texture, lighting, and crack morphology that stress-tests a method's boundary-alignment assumptions. Its comparatively small image resolution makes it the least demanding dataset in terms of computational scale, isolating the representation question from the scalability question.

\textit{SkyScapes}~\cite{azimi2019skyscapes} provides 16 aerial images at $5616\times3744$ (13\,cm/pixel GSD) with dense multi-class annotations over urban scenes. Lane markings span 1--3 pixels at this GSD and include both continuous and dashed patterns, the latter presenting a distinct challenge: each dash is a disconnected foreground component by design, so correct segmentation requires preserving many small isolated regions rather than a single connected structure. We evaluate on both the full 20-class dense task (SkyScapes Dense) and the 13-class lane-marking subtask (SkyScapes Lane). The dense task includes large-area classes---buildings, roads, vegetation---where pixel-level encoders with broad receptive fields hold an inherent advantage; including it provides an honest assessment of where a region-level representation loses ground. The small dataset size (16 images total) also tests whether SEMIR's few-shot parameter optimization and lightweight GNN generalize from minimal training data.

\begin{table}[t]
\centering
\caption{Dataset characteristics.}\label{tab:datasets}
\small
\begin{tabular}{lccccc}
\toprule
Dataset & Domain & Images & Resolution & Width (px) & Split \\
\midrule
TTPLA & Power lines & 1100 & $3840\!\times\!2160$ & 1--2 & official \\
CrackSeg9k & Cracks & 9255 & $400\!\times\!400$ & 1--3 & official \\
SkyScapes & Lane markings & 16 & $5616\!\times\!3744$ & 1--3 & official \\
\bottomrule
\end{tabular}
\end{table}

\subsection{Implementation Details}\label{sec:experiments:impl}

All experiments use an NVIDIA T4 GPU (16\,GB) with PyTorch. Minor parameters $\Theta$ are optimized via SMBO with an ExtraTrees surrogate on a few-shot subset ($|\mathcal{D}_{\mathrm{few}}| = 10$); optimized parameters are fixed for all subsequent training and evaluation. The GNN is a 3-layer GINE (hidden dimension 128, ReLU activation, batch normalization) trained with Adam ($\mathrm{lr}=10^{-3}$) for up to 200 epochs with early stopping on validation Dice (patience 10). All SEMIR scores report mean $\pm$ std over 5 independent runs. No dataset-specific architecture modifications, loss functions, or post-processing are applied: the same pipeline and GNN hyperparameters are used across all three benchmarks.

\subsection{Metrics}\label{sec:experiments:metrics}

We evaluate segmentation quality with four complementary metrics. \textbf{Dice coefficient} (F$_1$) and \textbf{Intersection-over-Union} (IoU) measure region overlap between prediction and ground truth; both range from 0 to~1, with Dice$\;= 2\,\text{IoU}/(1+\text{IoU})$. To assess boundary fidelity we report \textbf{Boundary~F$_1$} (BF1)~\cite{csurka2013good}, which computes precision and recall over predicted boundary pixels within a tolerance of 2\,px of the ground-truth boundary. Finally, we report \textbf{component fragmentation} $\Delta C = |C(\hat{Y}) - C(Y)|$, where $C(\cdot)$ counts the number of connected components in the foreground mask. A prediction that fragments a single crack into several disjoint segments, or merges two distinct power lines, incurs a large $\Delta C$ even if its pixel-level overlap is high. Dice and IoU are reported for all experiments; BF1 and $\Delta C$ are reported in the ablation study (Sec.~\ref{sec:experiments:ablation}), where we control both sides of the comparison and can compute them directly.

\subsection{Baselines}\label{sec:experiments:baselines}

We compare against two groups of methods.  The first consists of four widely adopted general-purpose segmentation architectures: UNet~\cite{ronneberger2015unet}, an encoder--decoder with skip connections that remains a standard reference across domains; DeepLabV3+~\cite{chen2018deeplabv3plus}, which combines atrous spatial pyramid pooling with a lightweight decoder for multi-scale context; PSPNet~\cite{zhao2017pspnet}, which aggregates global context through pyramid pooling; and SegFormer~\cite{xie2021segformer}, a hierarchical vision transformer with an MLP decode head. These four span the CNN-only, dilated-convolution, pooling-based, and transformer design families, providing broad coverage of mainstream encoder--decoder paradigms.

The second group comprises domain-specific architectures included because they hold leading published performance on individual benchmarks: PL-Deeplab~\cite{chen2023power} and PL-UNet~\cite{zhao2025pl} for power-line segmentation on TTPLA; CrackScopeNet~\cite{zhang2024crackscopenet} and BBCNet~\cite{hu2025bbcnet} for crack segmentation on CrackSeg9k; and SkyScapesNet~\cite{azimi2019skyscapes} for aerial scene parsing on SkyScapes. Each incorporates task-specific modules---edge-detection branches, frequency-domain enhancement, or multi-scale attention designed for thin structures---and is trained and evaluated on its target dataset only.  All baseline values are cited from their respective publications on official splits.

We report published numbers as-is; evaluation protocols may differ; our goal is cross-domain positioning rather than perfectly controlled reimplementation.

\subsection{Main Results}\label{sec:experiments:results}

\begin{table}[t]
\centering
\caption{Cross-domain positioning against published baselines on three thin-structure benchmarks. All baseline values cited from their respective source; SEMIR reports mean $\pm$ std over 5 runs on official splits. Best in \textbf{bold}, second best \underline{underlined}. $^\dagger$Domain-specific method.}
\label{tab:main}
\small
\begin{tabular*}{\linewidth}{@{\extracolsep{\fill}}l cc cc cc cc}
\toprule
 & \multicolumn{2}{c}{TTPLA} & \multicolumn{2}{c}{CrackSeg9k} & \multicolumn{2}{c}{SkyScapes} & \multicolumn{2}{c}{SkyScapes Lane} \\
\cmidrule(lr){2-3} \cmidrule(lr){4-5} \cmidrule(lr){6-7} \cmidrule(lr){8-9}
Method & Dice & IoU & Dice & IoU & Dice & IoU & Dice & IoU\\
\midrule

UNet
 & .841\tc{wei2024power}            & .726\tc{wei2024power}
 & .888\tc{zhang2024crackscopenet}  & .814\tc{zhang2024crackscopenet}
 & .248\tc{azimi2019skyscapes}      & .142\tc{azimi2019skyscapes} 
 & .165\tc{azimi2019skyscapes}      & .090\tc{azimi2019skyscapes}\\
DeepLabV3+
 & .833\tc{chen2023power}           & .715\tc{chen2023power}
 & .884\tc{zhang2024crackscopenet}  & .810\tc{zhang2024crackscopenet}
 & .553\tc{azimi2019skyscapes}      & .382\tc{azimi2019skyscapes} 
 & .542\tc{azimi2019skyscapes}      & .371\tc{azimi2019skyscapes}\\
PSPNet
 & .807\tc{zhao2025pl}              & .676 \tc{zhao2025pl}
 & .890\tc{zhang2024crackscopenet}  & .817\tc{zhang2024crackscopenet}
 & .467\tc{azimi2019skyscapes}      & .304\tc{azimi2019skyscapes} 
 & .528\tc{azimi2019skyscapes}      & .359\tc{azimi2019skyscapes}\\
SegFormer
 & .873\tc{wei2025unet}             & .776\tc{wei2025unet}
 & .889\tc{zhang2024crackscopenet}  & .816\tc{zhang2024crackscopenet}
 & ---                              & ---
 & .503\tc{liu2024advancements}     & .336\tc{liu2024advancements} \\
PL-Deeplab$^\dagger$
 & .877\tc{chen2023power}& .782\tc{chen2023power}
 & ---                              & ---
 & ---                              & --- 
 & ---                              & ---\\
PL-UNet 
& \underline{.889}\tc{zhao2025pl}       & \underline{.800}\tc{zhao2025pl}
 & ---                              & ---
 & ---                              & --- 
 & ---                              & ---\\
CrackScopeNet$^\dagger$
 & ---  & ---
 & {.893}\tc{zhang2024crackscopenet} & {.822}\tc{zhang2024crackscopenet}
 & ---  & --- 
 & ---                              & ---\\
 BBCNet
  & ---  & ---
 & \underline{.909}\tc{hu2025bbcnet}  & \underline{.833}\tc{hu2025bbcnet} 
 & ---  & ---
 & ---  & ---\\
  SkyScapesNet$^\dagger$
 & ---  & ---
 & ---  & ---
 & \textbf{.573}\tc{azimi2019skyscapes} & \textbf{.401}\tc{azimi2019skyscapes} 
 & \underline{.683}\tc{azimi2019skyscapes} & \underline{.519}\tc{azimi2019skyscapes} \\
\midrule
\textbf{SEMIR}
 & \textbf{.894} & \textbf{.808}
 & \textbf{.911} & \textbf{.837}
 & \underline{.562} & \underline{.391} 
 & \textbf{.713} & \textbf{.554}\\
\bottomrule
\end{tabular*}
\end{table}
Table~\ref{tab:main} compares SEMIR against published baselines on four benchmarks spanning three application domains.  All baseline values are cited from published sources; SEMIR numbers report the mean over five runs on official dataset splits.  The largest standard deviation across all reported metrics was $\pm .008$; per-run breakdowns are provided in the supplementary material but omitted here for space.

On \textbf{TTPLA}, SEMIR achieves .894 Dice and .808 IoU, surpassing PL-UNet (.889/.800), the previous best published result, as well as domain-specific PL-Deeplab (.877/.782) and general-purpose baselines including SegFormer (.873/.776).  Notably, SEMIR uses a single pipeline with no power-line-specific design choices, whereas PL-UNet and PL-Deeplab both incorporate domain-tailored modules.

On \textbf{CrackSeg9k}, SEMIR attains .911 Dice and .837 IoU, exceeding BBCNet (.909/.833) and CrackScopeNet (.893/.822). The general-purpose baselines cluster tightly between .884 and .890 Dice, suggesting that standard encoder--decoder architectures plateau on this dataset; SEMIR's graph-minor representation breaks through this ceiling.

On \textbf{SkyScapes Dense}, SkyScapesNet retains the top position (.573/.401) with SEMIR close behind (.562/.391).  This 20-class task includes large-area classes (buildings, roads, vegetation) where pixel-level encoders benefit from broad spatial context that a region-level GNN does not exploit.  The gap narrows considerably on \textbf{SkyScapes Lane}, the 13-class lane-marking subtask where all foreground classes are thin structures: SEMIR leads with .713 Dice and .554 IoU versus SkyScapesNet's .683/.519.  This confirms that the minor representation is most advantageous when the structures of interest are fine-grained and boundary-sensitive.

Across all four benchmarks, SEMIR achieves the best or second-best result using an identical pipeline---same GINE architecture, same few-shot SMBO tuning, same feature set---with no dataset-specific modifications.  The only method that outperforms SEMIR on any benchmark (SkyScapesNet on SkyScapes Dense) is a purpose-built multi-task architecture with edge-detection branches designed for that specific dataset.

\subsection{Ablation: Minor vs.\ SLIC}\label{sec:experiments:ablation}
\begin{table}[t]
\centering
\caption{Ablation: graph-minor partition vs.\ SLIC superpixels on CrackSeg9k. SLIC run at varying region counts with $c{=}10$, $\sigma{=}0$. The minor averages $\bar{n}$ regions per image (not constrained to match SLIC). $^*$SEMIR dynamically allocates regions per image.}
\label{tab:ablation_slic}
\small
\begin{tabular*}{\linewidth}{@{\extracolsep{\fill}}l r cccc}
\toprule
Partition & $\bar{n}$ & Dice & IoU & BF1 & $\Delta C$\,$\downarrow$ \\
\midrule
SLIC  & 25      & .414 & .261       & .190 & \underline{23.512} \\
SLIC  & 100     & .696 & .534       & .387 & 42.911 \\
SLIC  & 500     & .819 & .693       & .512 & 59.001 \\
SLIC  & 2000    & \underline{.851} & \underline{.741} & .696 & 73.122 \\
SLIC  & 10,000  & .791 & .654       & .578 & 80.090 \\
\midrule
Minor (SEMIR)   & 37.35$^*$ & \textbf{.907} & \textbf{.829} & \textbf{.736} & \textbf{5.103} \\
\bottomrule
\end{tabular*}
\end{table}
To isolate whether improvements derive from the topology-preserving minor or from region-level GNN inference generally, we replace the minor with SLIC superpixels~\cite{achanta2012slic} and retrain the same GINE classifier on the resulting region-adjacency graph.  All other components---features, architecture, training protocol---remain identical.  Rather than constraining SLIC to match the minor's region count (which would unfairly limit its operating regime), we sweep the number of segments across a wide range and report each configuration independently.  SLIC uses default compactness $c{=}10$ and no pre-smoothing ($\sigma{=}0$); the minor's region count is a per-image byproduct of the contraction process and averages $\bar{n}\approx37$ on CrackSeg9k.  Table~\ref{tab:ablation_slic} reports results.

\paragraph{Pixel overlap.}
SLIC's best Dice (.851 at $\bar{n}{=}2{,}000$) falls short of the minor's .907 by over five points, despite using roughly $50\times$ more regions.  Performance degrades at both extremes: too few superpixels under-segment the image, collapsing crack and background into shared regions (.414 Dice at $\bar{n}{=}25$), while too many superpixels fragment predictions and introduce classification noise (.791 at $\bar{n}{=}10{,}000$).  The minor avoids this trade-off entirely because the contraction process adapts region count to image content.

\paragraph{Boundary fidelity.}
The gap widens on BF1.  SLIC peaks at .696 ($\bar{n}{=}2{,}000$), while the minor reaches .736.  SLIC boundaries follow local color gradients, which correlate only loosely with thin-structure edges; the minor's boundaries are, by construction, aligned with the edges that survive contraction.

\paragraph{Topological consistency.}
The most striking contrast appears in $\Delta C$.  The minor achieves a component fragmentation of 5.1, meaning its predictions differ from the ground truth by roughly five connected components per image on average.  SLIC configurations range from 23.5 ($\bar{n}{=}25$, where heavy under-segmentation merges components) to 80.1 ($\bar{n}{=}10{,}000$, where over-segmentation shatters them).  Even the best SLIC operating point incurs nearly five times the fragmentation error of the minor.  This confirms that the minor's contraction-based construction preserves the connected-component structure of thin foreground regions in a way that color-based superpixels fundamentally cannot.

\paragraph{Parameter sensitivity.}
We sweep each minor parameter independently on CrackSeg9k while holding the remaining parameters at $\Theta_{\mathrm{opt}}$. Dice is relatively stable across a broad range of the contraction threshold $\psi$ but degrades sharply at both extremes: under-contraction ($\psi \ll \psi_{\mathrm{opt}}$) leaves the graph nearly unpruned, producing thousands of single-pixel supernodes that overwhelm the GNN, while over-contraction ($\psi \gg \psi_{\mathrm{opt}}$) merges crack and background into shared supernodes, collapsing foreground structure. $\Delta C$ is most sensitive to $\psi$, confirming that the contraction threshold is the primary lever for topology preservation. The edge deletion threshold $\alpha$ exhibits a gentler influence: values near $\alpha_{\mathrm{opt}}$ produce comparable Dice, with degradation only when $\alpha$ is set so low that nearly all edges are severed (isolating supernodes) or so high that no edges are removed (preventing boundary formation). Area bounds $\beta$ have the narrowest effect, filtering only extreme-size supernodes whose removal has limited impact on overlap or topology.  Across all three parameters, the few-shot SMBO consistently recovers values near the Pareto front of the Dice--$\Delta C$ trade-off, indicating that ten labeled images suffice for reliable parameter selection without manual intervention.

\subsection{Qualitative Results}\label{sec:experiments:qualitative}

\begin{figure}[ht]
    \centering
    \includegraphics[width=1\linewidth]{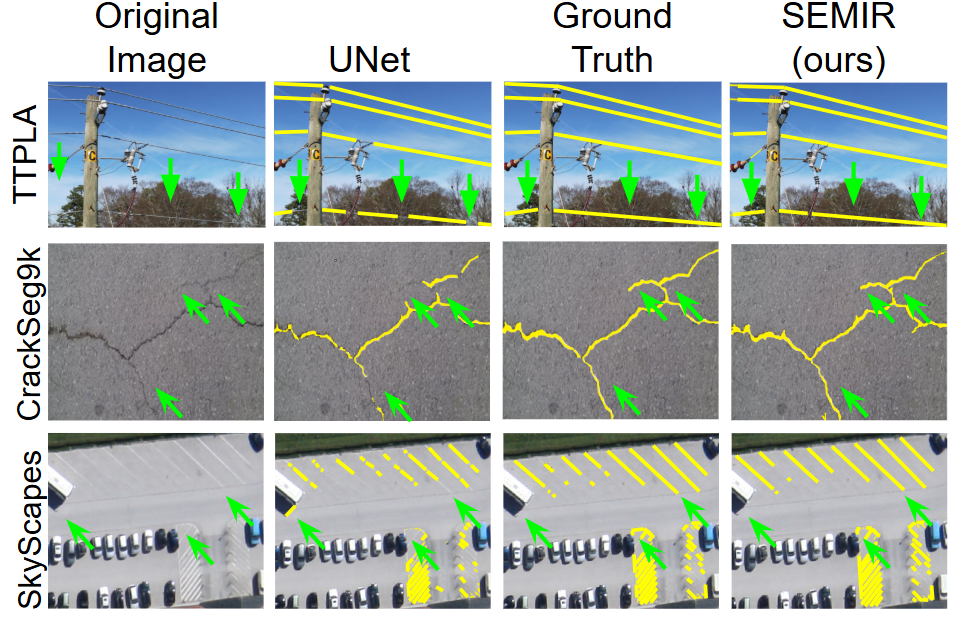}
    \caption{Qualitative comparison on one image per dataset. Foreground masks are overlaid in yellow with width dilated for visibility; green arrows highlight regions of interest. Columns: original image, UNet prediction, ground truth, SEMIR prediction. Rows: TTPLA (power lines), CrackSeg9k (pavement cracks), SkyScapes (lane markings).}
    \label{fig:qual}
\end{figure}

Figure~\ref{fig:qual} presents representative predictions from SEMIR alongside UNet on one image from each dataset. On TTPLA, UNet fragments continuous power lines where they cross cluttered background (green arrows); SEMIR preserves them as connected structures, consistent with its Dice advantage over general-purpose baselines reported in Table~\ref{tab:main}. On CrackSeg9k, UNet fragments cracks and introduces spurious branches near intersections, while SEMIR tracks the ground-truth topology more closely—producing fewer disconnected segments and fewer false positives at branch points. On SkyScapes, UNet misses faded parking markings that occupy only a few pixels in width; SEMIR recovers these, consistent with its IoU advantage on the lane-marking subtask. Across all three domains the primary distinction is topological rather than pixel-level: SEMIR produces fewer disconnected fragments and fewer false branches, matching the lower mask fragmentation reported in the ablation (Table~\ref{tab:ablation_slic}).

\subsection{Computational Complexity}\label{sec:experiments:complexity}

\paragraph{Minor construction.}
Each pixel is visited exactly once during the coprime traversal, and each edge is examined at most twice (once from each endpoint). Construction is therefore $O(N_c \cdot HW)$ where $N_c = 8$ is the grid connectivity---linear in pixel count regardless of the induced minor size. On TTPLA ($3840 \times 2160$, ${\sim}8.3 \times 10^6$ pixels), minor construction completes on CPU; no GPU is required for this stage.

\paragraph{GNN inference.}
Message passing operates on the minor $H$, with cost $O(L(|V(H)| + |E(H)|))$ for $L$ layers. The critical quantity is $|V(H)|$, which depends on image content rather than resolution: $|V(H)|$ reflects the number of intensity-homogeneous regions satisfying the contraction threshold $\psi$. Table~\ref{tab:complexity} reports empirical supernode counts. TTPLA images (${\sim}8.3 \times 10^6$ pixels) reduce to ${\sim}50$ supernodes on average---a reduction factor exceeding $10^5$. CrackSeg9k ($1.6 \times 10^5$ pixels) yields ${\sim}37$ supernodes; SkyScapes (${\sim}2.1 \times 10^7$ pixels) yields ${\sim}90$. In every case the GNN operates on a graph small enough to classify in a single forward pass with negligible memory, and the reduction grows with image resolution---precisely the regime where dense methods become most expensive.

\begin{table}[t]
\centering
\caption{Empirical supernode counts $|V(H)|$ and reduction factors across benchmarks. Supernode count depends on image content, not resolution: larger images with more homogeneous background yield greater reduction.}
\label{tab:complexity}
\small
\begin{tabular*}{\linewidth}{@{\extracolsep{\fill}}l c c c}
\toprule
& \textbf{TTPLA} & \textbf{CrackSeg9k} & \textbf{SkyScapes} \\
\midrule
Resolution & $3840 \times 2160$ & $400 \times 400$ & $5616 \times 3744$ \\
Pixels $|V(G)|$ & ${\sim}8.3 \times 10^6$ & $1.6 \times 10^5$ & ${\sim}2.1 \times 10^7$ \\
Supernodes $|V(H)|$ & $49.7 \pm 12.3$ & $37.4 \pm 8.1$ & $91.2 \pm 18.6$ \\
Reduction & ${\sim}1.7 \times 10^5 \!\times$ & ${\sim}4.3 \times 10^3 \!\times$ & ${\sim}2.3 \times 10^5 \!\times$ \\
\bottomrule
\end{tabular*}
\end{table}

\paragraph{Structure--complexity duality.}
Dense segmentation methods---UNet, DeepLabV3+, SegFormer---perform per-pixel inference, scaling directly with image resolution regardless of structural complexity. Patch-based inference reduces peak memory but not total computation, as overlapping windows collectively cover all pixels. SEMIR inverts this relationship: inference cost scales with the topological complexity of semantic boundaries rather than the resolution of the underlying lattice. This property is what enables full-resolution inference on images up to 21 megapixel scale without patching, as noted in the abstract. The duality becomes increasingly advantageous as acquisition resolution grows---exactly the trend in aerial, satellite, and infrastructure imaging where thin-structure segmentation is most needed.

\section{Limitations}\label{sec:limitations}

SEMIR's region-level representation trades spatial context breadth for boundary precision.  On tasks dominated by large-area classes, such as SkyScapes Dense, pixel-level encoders that aggregate features over wide receptive fields retain an advantage---SEMIR places second behind the purpose-built SkyScapesNet on this benchmark.  The minor contraction process also introduces a dependency on the quality of the initial oversegmentation: if early contraction steps merge a thin structure with its background, no downstream classification can recover it.  Finally, the current framework treats topology preservation purely at the representation level.  Combining the minor with loss functions that explicitly penalise topological errors---such as clDice~\cite{shit2021cldice} or skeleton-recall loss---could yield complementary gains, as these approaches target the optimisation objective rather than the input representation.  We leave this combination, along with extension to 3D volumetric structures, as future work.



\section{Conclusion}\label{sec:conclusion}

We presented SEMIR, a segmentation framework that replaces pixel-level inference with classification over a compact graph minor. By contracting homogeneous image regions into supernodes whose boundaries align with structural edges, SEMIR reduces the inference domain from millions of pixels to tens or hundreds of boundary-aligned regions while preserving the connected-component topology of thin foreground structures---this follows from the surjective structure of edge contraction, not from the training objective. A lightweight GINE classifier on the resulting region-adjacency graph, tuned via few-shot optimization on as few as ten images, achieves state-of-the-art results on TTPLA, CrackSeg9k, and SkyScapes Lane using an identical pipeline with no dataset-specific modifications to the GNN model. The controlled ablation against SLIC confirms that these gains originate in the topology-preserving representation itself, not in region-level inference generally.

More broadly, SEMIR demonstrates that the choice of inference representation---not only the architecture or loss---is a first-class design decision for structured prediction. The graph minor provides a principled middle ground between the pixel lattice (exact but redundant) and learned pooling (compact but lossy): it is compact, boundary-aligned, and equipped with a formal connectivity constraints and an exact lifting map. These properties are not specific to thin structures or to 2D images; extending the framework to 3D volumetric data, combining it with topology-aware losses that operate on the objective rather than the representation, and applying the minor to tasks beyond segmentation---such as structure-aware retrieval or graph-based active learning---are natural directions for future work.

\section*{Acknowledgements}
Luke Miller and Yugyung Lee acknowledge support from the National Science Foundation (NSF) under Award No. 2152057.

\bibliographystyle{splncs04}
\bibliography{ref}
\end{document}